\documentclass[conference]{IEEEtran}
\IEEEoverridecommandlockouts
\usepackage{cite}
\usepackage{placeins}
\usepackage{amsmath,amssymb,amsfonts}
\usepackage{algorithmic}
\usepackage{graphicx}
\usepackage{textcomp}
\usepackage{xcolor}
\def\BibTeX{{\rm B\kern-.05em{\sc i\kern-.025em b}\kern-.08em
    T\kern-.1667em\lower.7ex\hbox{E}\kern-.125emX}}
\begin{document}

\title{CheX-DS:  Improving Chest X-ray Image Classification with Ensemble Learning Based on DenseNet and Swin Transformer\\}

\author{\IEEEauthorblockN{Xinran Li}
\IEEEauthorblockA{\textit{School of Software Technology} \\
\textit{Dalian University of Technology}\\
Dalian, China \\
963707605@mail.dlut.edu.cn}
\\
\IEEEauthorblockN{Yu Liu}
\IEEEauthorblockA{\textit{School of Software Technology} \\
\textit{Dalian University of Technology}\\
Dalian, China \\
yuliu@dlut.edu.cn}
\and
\IEEEauthorblockN{Xiujuan Xu}
\IEEEauthorblockA{\textit{School of Software Technology} \\
\textit{Dalian University of Technology}\\
Dalian, China \\
xjxu@dlut.edu.cn}

\\
\IEEEauthorblockN{Xiaowei Zhao\textsuperscript{*}}
\IEEEauthorblockA{\textit{School of Software Technology} \\
\textit{Dalian University of Technology}\\
Dalian, China \\
xiaowei.zhao@dlut.edu.cn}
\thanks{\textit{*Corresponding author}}
}

\maketitle

\begin{abstract}
The automatic diagnosis of chest diseases is a popular and challenging task. Most current methods are based on convolutional neural networks (CNNs), which focus on local features while neglecting global features. Recently, self-attention mechanisms have been introduced into the field of computer vision, demonstrating superior performance. Therefore, this paper proposes an effective model, CheX-DS, for classifying long-tail multi-label data in the medical field of chest X-rays. The model is based on the excellent CNN model DenseNet for medical imaging and the newly popular Swin Transformer model, utilizing ensemble deep learning techniques to combine the two models and leverage the advantages of both CNNs and Transformers. The loss function of CheX-DS combines weighted binary cross-entropy loss with asymmetric loss, effectively addressing the issue of data imbalance. The NIH ChestX-ray14 dataset is selected to evaluate the model's effectiveness. The model outperforms previous studies with an excellent average AUC score of 83.76\%, demonstrating its superior performance.
\end{abstract}


\section{Introduction}
In recent years, with the rapid development of deep learning technology, its application in the field of medical imaging has become increasingly widespread. Medical imaging, as an important medical discipline, plays a crucial role in diagnosing and treating various diseases. Among medical images, chest X-rays are one of the most common and play an irreplaceable role in the diagnosis of pulmonary diseases\cite{b1}. However, traditional diagnosis of chest X-rays relies on doctors' extensive experience and professional knowledge, which presents issues such as slow diagnostic speed and high subjectivity.

With the rapid development of artificial intelligence technology, deep learning has shown great potential and application value in medical image analysis\cite{b2}. However, there are three major challenges in the diagnosis of diseases from chest X-rays\cite{b3}: 

(1) Most current models are based on a single architecture, such as CNNs or ViTs, which have relatively weak performance. 

(2) The long-tail distribution of medical diseases. 

(3) The multi-label nature of medical diseases.

To date, deep learning technologies in the field of computer vision can be broadly categorized into two main classes: Convolutional Neural Networks (CNNs)\cite{b4}, and Vision Transformers (ViTs)\cite{b5}. Models for automated medical image diagnosis are mostly based on improved CNN algorithms, such as DenseNet\cite{b6}. CNNs' greatest advantage lies in their ability to automatically and efficiently extract multi-level features of images through local receptive fields and weight-sharing mechanisms, thereby achieving accurate image recognition and processing. However, CNN methods focus solely on local image features, neglecting global features\cite{b7}. ViTs introduce a multi-head self-attention mechanism, providing context-aware long-term dependencies and emphasizing more important global features\cite{b8}. The majority of methods consider only a single model and do not take advantage of ensemble learning techniques to integrate the strengths of multiple models\cite{b9}.

The presence of numerous rare diseases contributes to a significant long-tail distribution in the dataset for pulmonary diseases, where a minority of classes (head classes) constitute a vast proportion, while the majority of classes (tail classes) occupy a minimal proportion\cite{b10}. For example, there are numerous cases of Infiltration, while Hernia cases are relatively scarce. Models trained on such data tend to favor head classes, neglecting tail classes. Chest X-ray data, being a typical example of long-tail distribution, poses immense challenges for models trained on it.

Pulmonary diseases often coexist with multiple complications, making image data typically exhibit multi-label properties\cite{b11}. Existing models often fail to fully address multi-label issues, leading to inaccurate disease diagnoses. For instance, a patient may be diagnosed with concurrent conditions such as cardiomegaly and edema.

This paper introduces an effective long-tail multi-label data classification model, CheX-DS, for chest X-ray images in the medical domain. The model is based on the commonly used convolutional neural network model, DenseNet\cite{b12}, and the popular and excellent Swin Transformer\cite{b13}, leveraging ensemble learning techniques to integrate the strengths of both. CheX-DS utilizes pre-training on the NIH ChestX-ray14 dataset to enhance the model's ability to learn different disease features\cite{b14}. Given the inherent imbalance in multi-label long-tail classification, CheX-DS's loss function combines weighted binary cross-entropy loss with asymmetric loss, addressing both inter-class and intra-class imbalances\cite{b3}. Through comparison with other existing methods, our approach demonstrates superior performance in terms of AUC, validating the effectiveness of our proposed method. Our main contributions are summarized as follows:

1. We proposed the ensemble model CheX-DS, based on DenseNet and Swin Transformer, which achieves better performance compared to individual models through ensemble learning techniques.

2. Through extensive experimentation, our model demonstrates superior performance compared to other existing methods.

3. The model effectively handles long-tail data by employing an enhanced loss function.

The structure of the paper is as follows: Section II discusses the application of Convolutional Neural Network (CNN) techniques and Transformer techniques in chest X-ray (CXR) analysis. The specific methods used in the paper are introduced in Section III. Experimental results are presented in Section IV. Section V concludes the paper.

\section{Related Work}
In recent years, with the advancements in deep learning technology, many methods have demonstrated superior performance in the field of medical imaging. In this section, we briefly summarize some deep learning methods used for CXR image analysis.

\subsection{Convolutional Neural Networks in medical image domain}
Classification of chest X-rays (CXRs) is a typical multi-label classification task. In the early stages, methods such as multi-label \textit{k}-nearest neighbors (MLkNN) were used to handle multi-label classification\cite{b15}. With the advancement of deep learning technology, convolutional neural network (CNN) techniques began to be applied to CXR classification. CNN methods such as AlexNet and VGGNet were used by Wang et al. to predict 14 diseases in the NIH ChestX-ray14 dataset\cite{b14}. Rajpurkar et al. designed CheXNet based on DenseNet121\cite{b6}. DenseNet is characterized by its dense connectivity, where each layer receives input from all preceding layers, improving feature reuse and reducing the number of parameters. Before the advent of Transformers, DenseNet was the relatively most effective method in the field of medical imaging. The architecture diagram of DenseNet121 is shown in Fig. 1.

\begin{figure}
    \centering
    \includegraphics[width=1\linewidth]{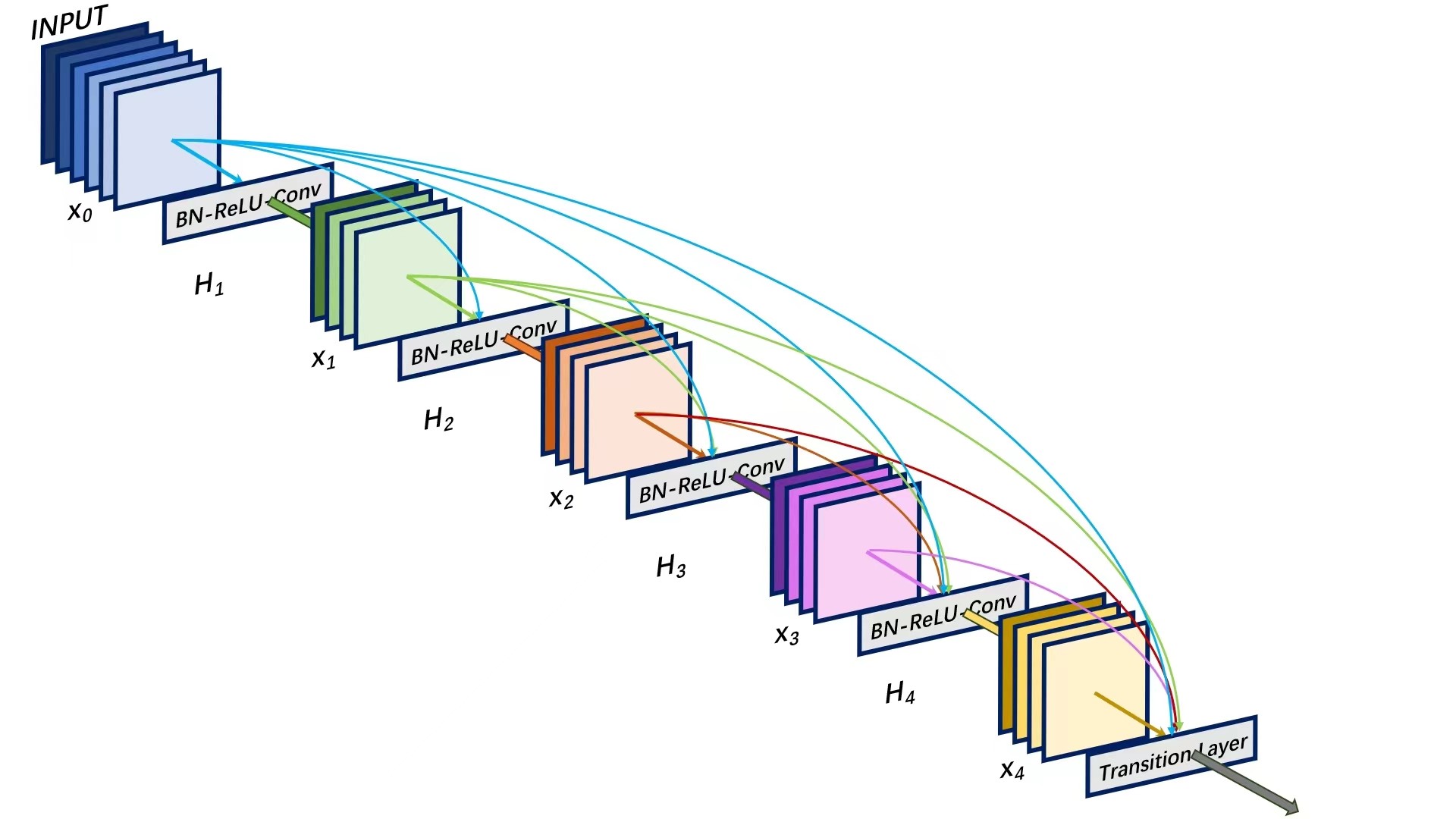}
    \caption{The architecture diagram of DenseNet121}
    \label{fig:enter-label}
\end{figure}
\begin{figure*}
    \centering
    \includegraphics[width=0.8\linewidth]{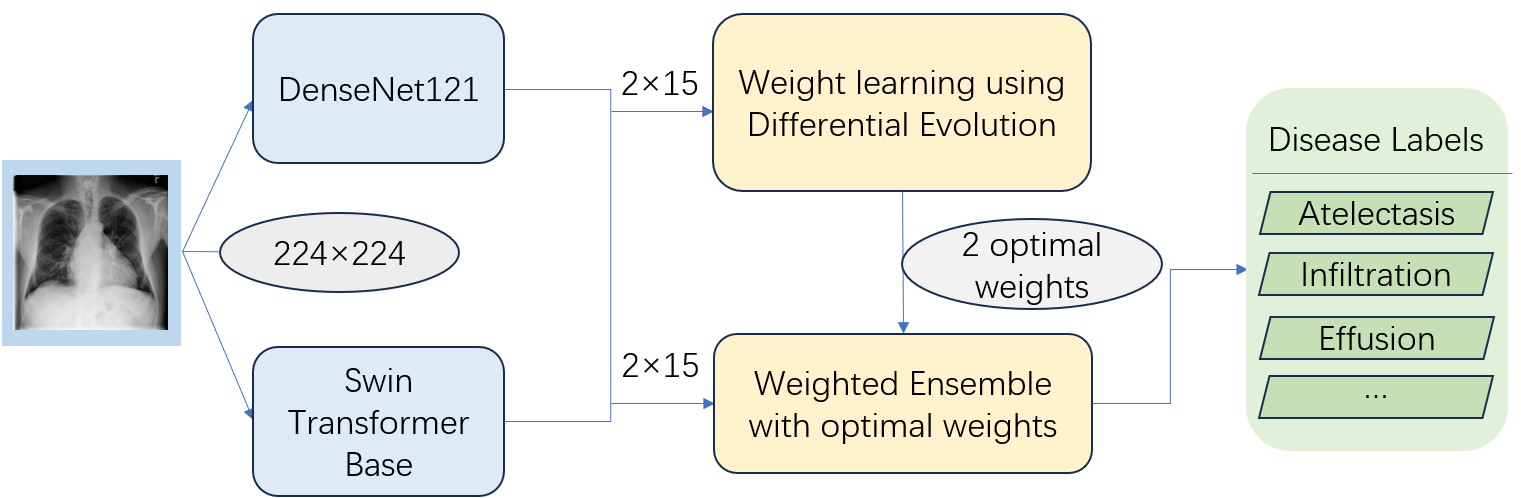}
    \caption{The schematic diagram of CheX-DS, featuring the average weighted ensemble with differential evolution.}
    \label{fig:enter-label}
\end{figure*}

The core idea of DenseNet is `dense connections', meaning that the output of each layer is connected to the outputs of all preceding layers. This design enables each layer to directly access the feature maps from all preceding layers, facilitating more direct and efficient information flow. Although CNN methods achieve good results, they tend to overlook global features and focus only on local features.

\subsection{Transformer in medical image domain}
With the introduction of Transformers, more and more people started applying them to computer vision tasks. Vision Transformer and Swin Transformer are both classic improved methods\cite{b13}\cite{b16}.

 Most studies combine CNNs with Transformers to further enhance the advantages of Transformers. Muhamad Faisal et al. fused CheXNet and ViT to propose the CheXViT model\cite{b17}. CheXViT combines the strengths of CNNs and Transformers, achieving superior performance in multi-label CXR image classification by leveraging CNNs' inductive biases and Transformers' ability to capture long-range feature dependencies. Manzari proposed MedViT, which combines the locality of CNNs with the global connectivity of vision Transformers to enhance robustness and efficiency in medical image diagnosis, particularly against adversarial attacks\cite{b18}. Dongkyun Kim proposed the fusion module CheXFusion based on Transformers for CXR, which achieved better results by improving the loss function\cite{b3}. Almalik proposed the SEViT model, which leverages the intermediate feature representations learned by the initial ViT model, and combines predictions from multiple classifiers based on these representations to enhance defense against adversarial attacks\cite{b19}.

Although Transformers can account for global features, their self-attention mechanism has a high computational complexity, especially when processing high-resolution images, resulting in rapidly increasing computational and memory requirements. Moreover, Transformers require a large amount of data for training. If the data volume is insufficient, their performance may not be as good as CNN.
 
Additionally, some studies have proposed the use of ensemble learning techniques, combining multiple base models to effectively enhance performance. Ashraf et al. proposed SynthEnsemble, which utilized ensemble deep learning to combine different models, achieving the best performance on the NIH ChestX-ray14 dataset\cite{b9}. This ensemble learning method can effectively combine the advantages of multiple models to form a better overall model. Therefore, this paper also utilize ensemble learning methods to effectively combine CNN and Transformer approaches.

\section{Method}
In this section, we introduce the methods employed in our study. Firstly, we present the two models utilized, DenseNet and Swin Transformer. Subsequently, we discuss the improved loss function employed. Finally, we outline the ensemble learning approach utilized. The schematic diagram of CheX-DS, featuring the average weighted ensemble with differential evolution, is illustrated in Fig. 2.

\subsection{DenseNet}\label{AA}
DenseNet (Densely Connected Convolutional Networks) is a deep learning architecture aimed at addressing the vanishing gradient problem and parameter efficiency in training deep neural networks. Proposed by Huang et al. in 2017, it's a variant of convolutional neural networks (CNNs). DenseNet121 is a specific variant of DenseNet, where the number `121' denotes the number of layers in the network. It consists of 121 layers, including four dense blocks. CheXNet, which employs DenseNet121, has achieved excellent results. 

\subsection{Swin Transformer}
Swin Transformer is a novel neural network model based on the Transformer architecture, proposed by Microsoft Research Asia. In contrast to traditional Transformer models, Swin Transformer introduces a novel visual perception mechanism, employing hierarchical attention mechanisms and block-based visual processing to handle large-scale images. It excels in processing large-sized images and has achieved state-of-the-art performance in many computer vision tasks. The structure of Swin Transformer is shown in Fig. 3.
\begin{figure*}
    \centering
    \includegraphics[width=0.75\linewidth]{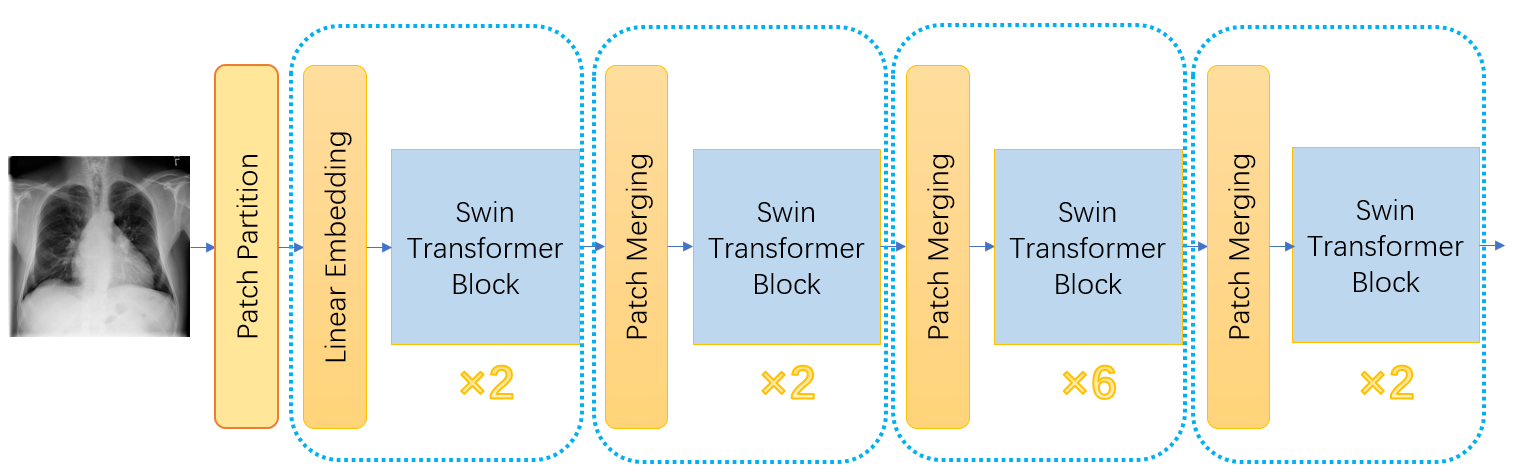}
    \caption{The architecture diagram of Swin Transformer}
    \label{fig:enter-label}
\end{figure*}
\subsection{Loss Function}
In multi-label classification, the commonly used loss function is binary cross-entropy loss. In multi-label long-tail classification, there are inter-class and intra-class imbalances. 

Inter-class imbalance refers to the unequal distribution of sample quantities among different categories in the dataset. To address inter-class imbalance, a weighted binary cross-entropy loss is utilized\cite{b20}:
\begin{equation}
L_{wbce}=-\sum_{i=1}^{C}w_i(y_ilog(p_i)+(1-y_i)log(1-p_i))
\end{equation}
where \(C\) is the total number of classes, and \(y_i\), \(p_i\), and \(w_i\) are ground truth labels, predicted probability, and weight for class \(i\). \(w_i=y_ie^{1-\rho}+(1-y_i)e^\rho\) where \(\rho\) is the ratio of positive samples for class \(i\).

Intra-class imbalance refers to the uneven distribution of different samples within the same category, such as having far more negative samples than positive samples. Asymmetric loss functions are commonly employed to address intra-class imbalance. One popular asymmetric loss function is a variant of Focal Loss. It adjusts the focal parameter \(\gamma\) to control the model's attention towards different categories\cite{b21}:
\begin{equation}
L_{asl}=-\sum_{i=1}^{C}((1-p_i)^{\gamma_+}y_ilog(p_i)+p_{mi}^{\gamma_-}(1-y_i)log(1-p_{mi}))
\end{equation}
where \(p_{mi}=max(p_i-m,0)\).

By combining the weighted binary cross-entropy loss with an asymmetric loss, we can effectively address both inter-class and intra-class imbalances in multi-label long-tail classification tasks. The improved loss function we use is as follows\cite{b3}:
\begin{equation}
L=-\sum_{i=1}^{C}w_i((1-p_i)^{\gamma_+}y_ilog(p_i)+p_{mi}^{\gamma_-}(1-y_i)log(1-p_{mi}))
\end{equation}

\subsection{Ensemble Learning}
Ensemble learning enhances overall performance by combining predictions from multiple models. We employed an ensemble method utilizing weighted averaging optimized through differential evolution\cite{b9}.

Each of the two models generates a probability vector for every image, indicating the predicted probabilities for each class. To derive the final prediction probability vector for each image, we employ different weights to average these individual probability vectors. These weights represent the contribution of each model to the final prediction.

We utilized a stochastic global search algorithm called differential evolution to determine the optimal weights for each model\cite{b22}. These weights, constrained to sum up to 1, ensure that the weighted average remains a valid probability distribution suitable for final predictions.

\section{Experiments}
In this section, we present the experimental details and results. Firstly, we introduce the NIH ChestX-ray14 dataset utilized in the experiments. Subsequently, we outline the experimental parameter settings and model evaluation parameters. Then, we compare the ensemble model with the two individual models that constitute it, highlighting the advantages of ensemble learning. Following that, we conduct a comparison of loss functions to underscore the effectiveness of the improved loss function. Finally, we compare the CheX-DS model with other existing models to demonstrate its superiority.

\subsection{Dataset}
Our dataset is the NIH ChestX-ray14, provided by the National Institutes of Health (NIH). It is currently one of the most widely used medical imaging datasets, extensively employed in various studies related to lung disease classification and diagnosis\cite{b14}.

\begin{table*}[]
\caption{The specific names of the 14 diseases and their respective proportions}
\resizebox{\textwidth}{!}{%
\begin{tabular}{l|llllllllllllll}
\hline
Disease & Atelectasis & Consolidation & Infiltration & Pneumothorax & Edema & Emphysema & Fibrosis & Effusion & Pneumonia & Pleural Thickening & Cardiomegaly & Nodule & Mass & Hernia \\ \hline
Proportion &22.33\% &9.02\% &38.44\% &10.24\% &4.45\% &4.86\% &3.26\% &25.73\% &2.76\% &6.54\% &5.36\% &12.23\% &11.17\% &0.44\%  \\ \hline
\end{tabular}%
}
\end{table*}

The NIH ChestX-ray14 dataset comprises 112,120 anterior-posterior (PA view) X-ray images collected from 30,805 patients. Each X-ray image is annotated by professional radiologists and is associated with 14 labels corresponding to various lung diseases. The specific names of the 14 diseases and their respective proportions are listed in Table I. Additionally, the dataset includes a `No Findings' category, used to label X-ray images where no diseases are detected. As shown in Fig. 4, the number of different diseases varies significantly. A few diseases have a large number of cases, while most diseases have relatively few cases, forming a typical long-tail distribution.

\begin{figure}
    \centering
    \includegraphics[width=1\linewidth]{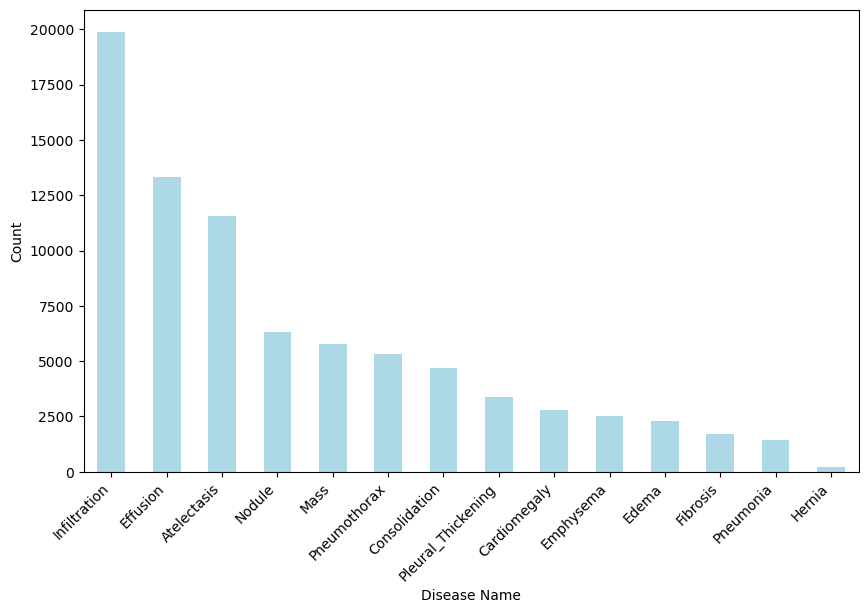}
    \caption{The distribution of the number of cases for the 14 diseases}
    \label{fig:enter-label}
\end{figure}

\subsection{Experimental Setup}
We fine-tuned DenseNet121 and Swin Transformer Base, which were pre-trained on ImageNet. The image size was adjusted from 1024\(\times\)1024 to 224\(\times\)224 pixels. We applied random horizontal flipping and rotation to the images, with an augmentation probability of 50\% and a rotation limit of ten degrees. The dataset was divided into three groups: 70\% for training, 20\% for testing, and 10\% for validation. In the loss function of the model, we set \(\gamma_+=1\), \(\gamma_-=4\), \(m=0.05\).

We have chosen PyTorch as the implementation platform. The experiments are run on an NVIDIA GeForce RTX 4090 D. We have opted to improve the loss function for multi-label classification. We are using AdamW as our optimizer, with a weight decay of 1e-2 for each DNN and momentum set to 0.9. The batch size is set to 64. The training process incorporates an early stopping mechanism.

\subsection{Evaluation Metrics}
According to previous CXR classification work, we evaluated our method using a commonly used metric in multi-label classification tasks:  Area Under the Receiver Operating Characteristic Curve (AUC-ROC). It measures the area under the curve plotted by the true positive rate (sensitivity) against the false positive rate (1\(-\)specificity) for different classification thresholds. An AUC value of 1 indicates perfect classification ability.

\subsection{Ensemble Model CheX-DS vs. Individual Models}
We compare the ensemble model CheX-DS with the two models that constitute it, DenseNet121 and Swin Transformer Base. All models were trained with the same parameters, using the same enhanced loss function, and fine-tuned on the same dataset. We computed the AUC scores for each class. Additionally, we calculated the average AUC score for all pathological combinations.

The AUC results for DenseNet, Swin Transformer, and CheX-DS with 15 classes are shown in Fig. 5, Fig. 6, and Fig. 7, respectively. The comparison of AUC for the three models across 15 classes is shown in TABLE II.

By examining the images and tables, it's evident that CheX-DS achieved the best AUC for the all of diseases, with notably higher average AUC compared to the two individual models. This indicates that ensemble learning significantly enhances performance.

\begin{table*}[t]
\caption{The AUC scores comparison between CheX-DS with 15 classes and the improved loss DenseNet and Swin Transformer}
\resizebox{\textwidth}{!}{%
\begin{tabular}{c|ccccccccccccccc|c}

\hline
Method   & Atel            & Cons            & Infi            & Pneumothorax    & Edem            & Emph            & Fibr            & Effu            & Pneumonia       & P\_T            & Card            & Nodu            & Mass            & Hern            & No\_Find        & Mean            \\ \hline
DenseNet & 0.8083          & 0.7886          & 0.7134          & 0.9006          & 0.8882          & 0.8986          & 0.8069          & 0.8849          & 0.7709          & 0.8041          & 0.8903          & 0.7641          & 0.8452          & 0.8794          & 0.7823          & 0.8284          \\
Swin\_B  & 0.8103          & 0.8017          & 0.7161 & 0.8943          & 0.8868          & 0.9033 & 0.7875          & 0.8815          & 0.7777          & 0.8004          & 0.8915          & 0.7479          & 0.8366          & 0.8584          & 0.7834          & 0.8251          \\
CheX-DS  & \textbf{0.8179} & \textbf{0.8030} & \textbf{0.7209}          & \textbf{0.9066} & \textbf{0.8942} & \textbf{0.9078}          & \textbf{0.8091} & \textbf{0.8880} & \textbf{0.7827} & \textbf{0.8118} & \textbf{0.9008} & \textbf{0.7681} & \textbf{0.8561} & \textbf{0.9081} & \textbf{0.7893} & \textbf{0.8376} \\ \hline
\end{tabular}%
}
\vspace{0.5cm} 

\caption{The comparison of AUC for ensemble models using BCE loss and the improved loss function across 15 classes}
\resizebox{\textwidth}{!}{%
\begin{tabular}{c|ccccccccccccccc|c}
\hline
Loss Function & Atel            & Cons            & Infi            & Pneumothorax    & Edem            & Emph            & Fibr            & Effu            & Pneumonia       & P\_T            & Card            & Nodu            & Mass            & Hern            & No\_Find        & Mean            \\ \hline
BCELoss       & \textbf{0.8190} & 0.7991          & 0.7200          & \textbf{0.9071} & \textbf{0.8961} & \textbf{0.9151} & 0.7947          & 0.8876          & \textbf{0.7862} & \textbf{0.8138} & 0.8982          & 0.7673          & 0.8506          & 0.8697          & 0.7891          & 0.8342          \\
Improved Loss & 0.8179          & \textbf{0.8030} & \textbf{0.7209} & 0.9066          & 0.8942          & 0.9078          & \textbf{0.8091} & \textbf{0.8880} & 0.7827          & 0.8118          & \textbf{0.9008} & \textbf{0.7681} & \textbf{0.8561} & \textbf{0.9081} & \textbf{0.7893} & \textbf{0.8376} \\ \hline
\end{tabular}%
}
\end{table*}
\vspace{0.5cm} 
\begin{table*}
\caption{The comparison of AUC for CheX-DS with other State-of-the-Art Benchmarks across 15 classes}
\resizebox{\textwidth}{!}{%
\begin{tabular}{c|ccccccccccccccc|c}
\hline
Method    & Atel           & Cons           & Infi           & Pneumothorax   & Edem           & Emph           & Fibr           & Effu           & Pneumonia      & P\_T           & Card           & Nodu           & Mass           & Hern           & No\_Find       & Mean           \\ \hline
CheXNet   & 0.769          & 0.745          & 0.694          & 0.852          & 0.842          & 0.906          & 0.821          & 0.825          & 0.715          & 0.766          & 0.885          & 0.759          & 0.824          & 0.901          & -              & 0.807          \\
DualCheXN & 0.784          & 0.746          & 0.705          & 0.876          & 0.852          & 0.942          & 0.837          & 0.831          & 0.727          & 0.796          & 0.888          & 0.796          & 0.838          & 0.912          & -              & 0.823          \\
CheXGCN   & 0.786          & 0.751          & 0.699          & 0.876          & 0.850          & \textbf{0.944} & 0.834          & 0.832          & 0.739          & 0.795          & 0.893          & \textbf{0.800} & 0.840          & 0.929          & -              & 0.826          \\
ImageGCN  & 0.802          & 0.796          & 0.702          & 0.900          & 0.883          & 0.915          & 0.825          & 0.874          & 0.715          & 0.791          & 0.894          & 0.768          & 0.843          & \textbf{0.943} & -              & 0.832          \\
CheXViT   & 0.807          & 0.785          & \textbf{0.724} & \textbf{0.911} & 0.873          & 0.935          & \textbf{0.849} & 0.860          & 0.756          & 0.807          & \textbf{0.924} & 0.792          & \textbf{0.877} & 0.905          & 0.760          & \textbf{0.838} \\
CheX-DS   & \textbf{0.818} & \textbf{0.803} & 0.721          & 0.907          & \textbf{0.894} & 0.908          & 0.809          & \textbf{0.888} & \textbf{0.783} & \textbf{0.812} & 0.901          & 0.768          & 0.856          & 0.908          & \textbf{0.789} & \textbf{0.838} \\ \hline
\end{tabular}%
}
\end{table*}

\begin{figure}
    \centering
    \includegraphics[width=1\linewidth]{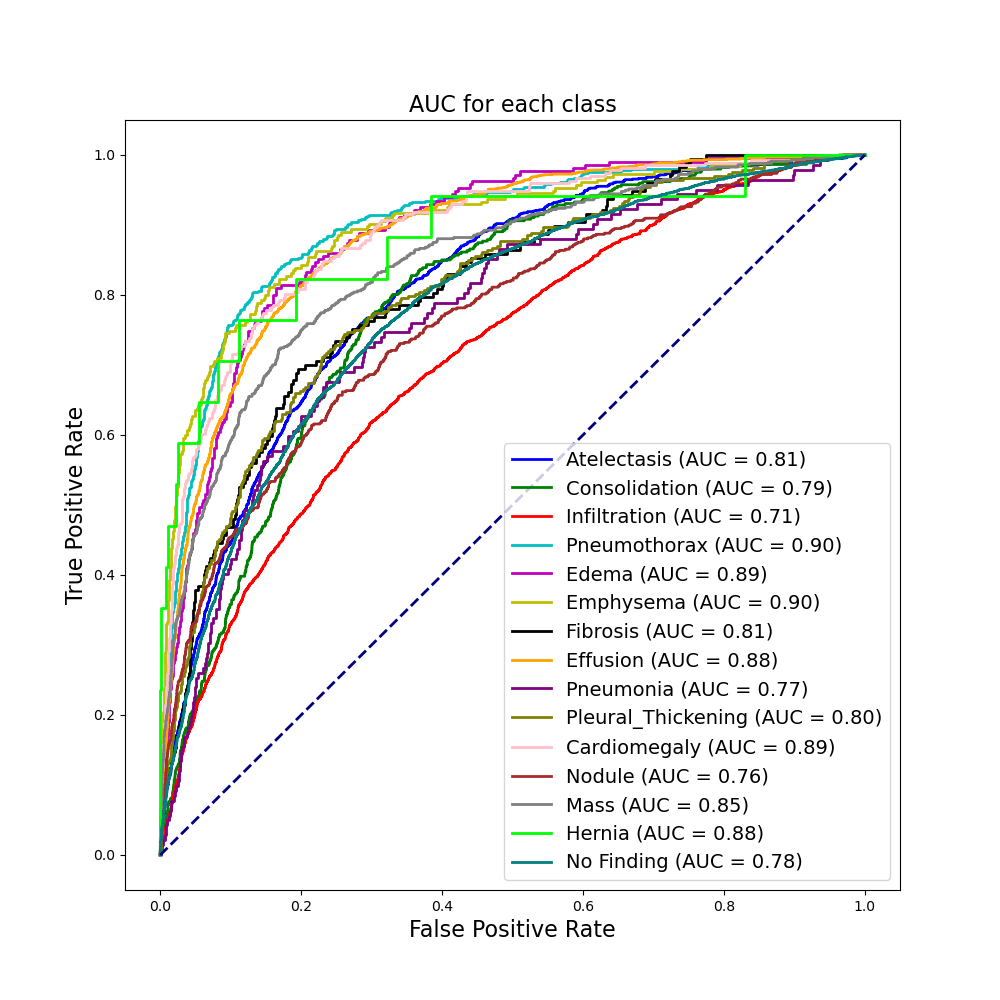}
    \caption{ROC Curves of improved loss DenseNet with 15 classes}
    \label{fig:enter-label}
\end{figure}
\begin{figure}
        \centering
        \includegraphics[width=1\linewidth]{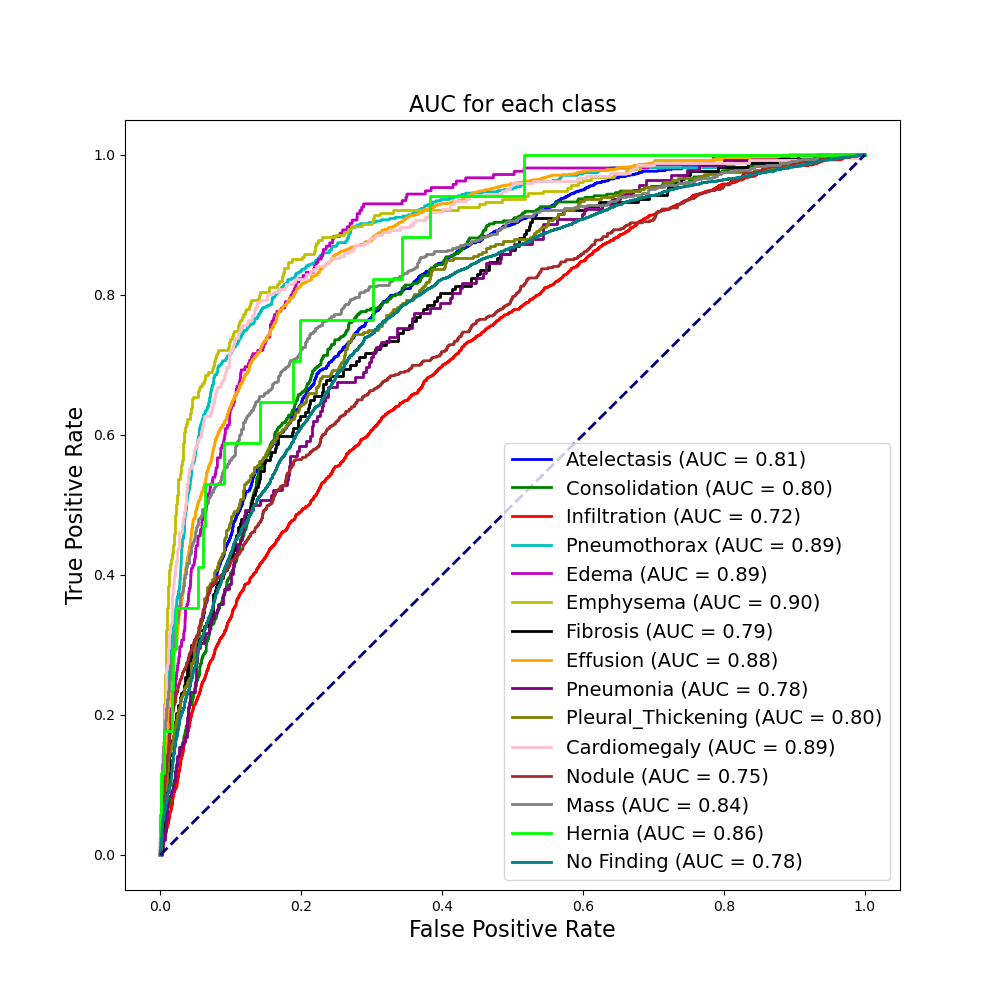}
        \caption{ROC Curves of improved loss Swin Transformer with 15 classes}
        \label{fig:enter-label}
\end{figure}
\begin{figure}
        \centering
        \includegraphics[width=1\linewidth]{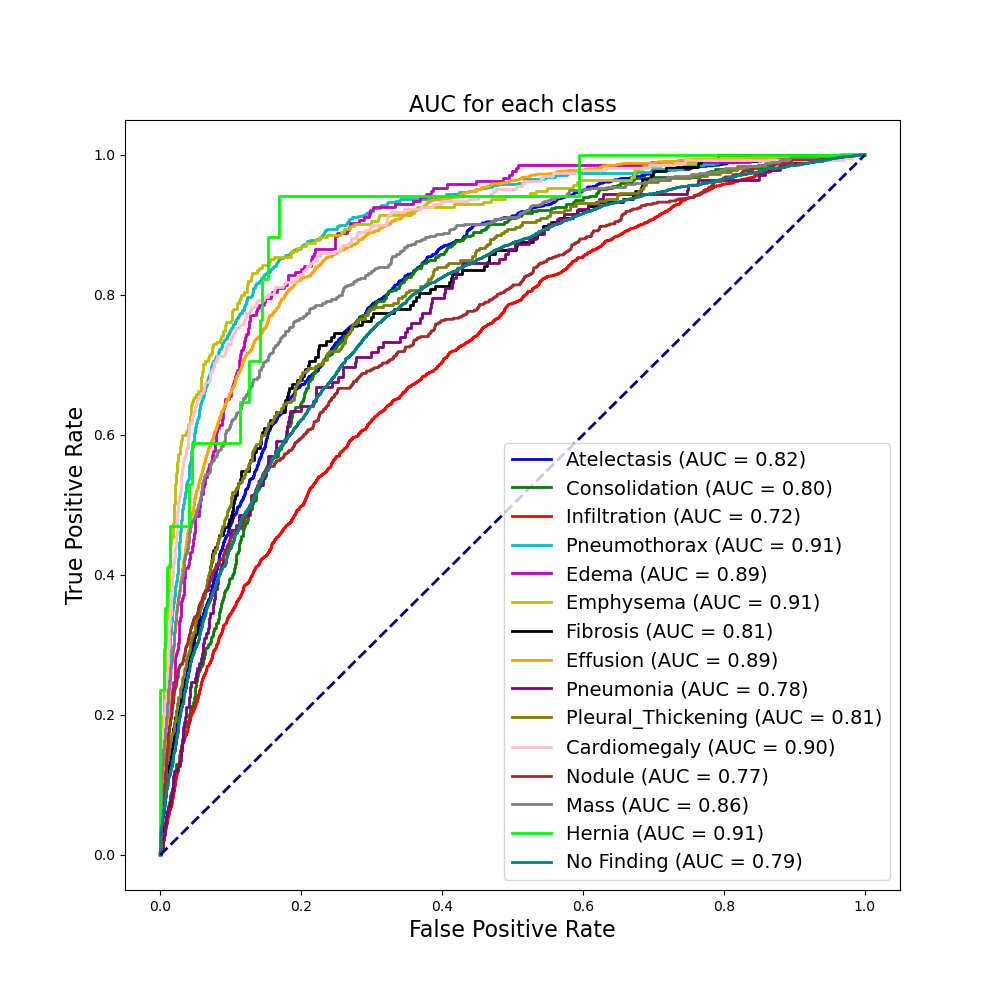}
        \caption{ROC Curves of CheX-DS with 15 classes}
        \label{fig:enter-label}
\end{figure}

\subsection{Comparison of loss functions}
To validate the superiority of the improved loss function, we trained and fine-tuned DenseNet and Swin Transformer using binary cross-entropy (BCE) loss functions. Consequently, we utilized ensemble learning to create an ensemble model using BCE loss. TABLE III presents the comparison of ensemble models using different loss functions.

By comparison, it can be observed that the ensemble model using the improved loss function achieves an average AUC score 0.003 higher than the ensemble model using BCELoss. This indicates that the improved loss function contributes to a certain improvement in performance.

\subsection{Comparison with existing approaches}
We compared CheX-DS with previous studies focusing on AUC to validate our research findings. The previous studies discussed in the comparison include CheXNet\cite{b6}, DualCheXN\cite{b23}, CheXGCN\cite{b24}, ImageGCN\cite{b25} and CheXViT\cite{b17}. The results are as shown in TABLE IV.

Through comparison, we can see that the CheX-DS model achieves the best performance on 7 disease indicators and the average indicator. What sets our model apart from other studies is that we also calculated the `No Finding' label, which was previously only done by CheXViT, and our model outperformed CheXViT on this label. Compared to other labels, the frequency of this label is higher, resulting in a significant imbalance between this label and all other labels, which may weaken the average AUC. However, our model outperforms other models.

\section{Conclusion}
In this paper, we constructed a multi-label classification model named CheX-DS for chest X-ray (CXR) images based on DenseNet121 and Swin Transformer Base using ensemble learning. Firstly, we fine-tuned pre-trained DenseNet and Swin Transformer on the NIH ChestX-ray14 dataset. We employed a loss function that combines weighted binary cross-entropy loss and asymmetric loss to address the long-tail distribution issue in the dataset. Subsequently, we used weighted averaging ensemble to combine the two models, with ensemble weights determined through differential evolution. Then, we compared and concluded that the ensemble model using the improved loss function outperformed the model using binary cross-entropy loss. Finally, through comparison with other existing models, our model achieved excellent performance with an average AUROC of 83.76\%. 

In future work, we will continue to utilize ensemble learning combined with improved base models to enhance performance. 

\vspace{12pt}
\color{red}
\end{document}